\documentclass[a4paper,10pt,conference]{IEEEtran}
\usepackage{array}
\usepackage{textcomp}
\usepackage{stfloats}
\usepackage{url}
\usepackage{verbatim}
\usepackage{graphicx}
\usepackage{cite}
\usepackage{amsmath,amssymb,amsfonts}
\usepackage{amsmath}
\usepackage{graphicx}
\usepackage{algorithm}
\usepackage{algorithmic}
\usepackage{subcaption}
\usepackage[colorinlistoftodos]{todonotes}
\usepackage{float}
\usepackage[colorlinks=false, allcolors=blue]{hyperref}
\usepackage{enumerate}
\usepackage{graphicx}
\usepackage{textcomp}
\usepackage{pdfpages}
\usepackage{multirow}
\usepackage{array}
\usepackage{pgfplots}
\usepackage{xcolor}
\usepackage{soul}
\usepackage{tikz}
\usepackage{pgfkeys}

\usepackage{amssymb}

\usepackage{graphicx}

\usepackage{multirow}
\usepackage{textcomp}

\usepackage{bbm}

\usepackage{pgfplots}
\usepgfplotslibrary{colorbrewer}
\pgfplotsset{compat=newest}
\usepackage{xcolor} 
\usepackage{caption} 

\begin{document}

\title{Enhancing Encrypted Internet Traffic Classification Through Advanced Data Augmentation Techniques}

\author
{
    \IEEEauthorblockN{Yehonatan Zion, Porat Aharon,  Ran Dubin\\}
    \IEEEauthorblockA{\textit{Department of Computer Science} \\ \textit{Ariel Cyber Innovation Center}\\ Ariel University, Israel\\
    yehonata.zion@msmail.ariel.ac.il; porataharon@gmail.com; rand@ariel.ac.il}
\and
    \IEEEauthorblockN{Amit Dvir\\}
    \IEEEauthorblockA{\textit{Dept. of Computer and Software Eng.}\\\textit{Ariel Cyber Innovation Center}\\ Ariel University, Israel\\
    amitdv@ariel.ac.il}
\and 
    \IEEEauthorblockN{Chen Hajaj\\}
    \IEEEauthorblockA{\textit{Department of Industrial Engineering and Management} \\\textit{Data Science and Artificial Intelligence Research Center}\\ Ariel University, Israel\\
    chenha@ariel.ac.il}
}


\maketitle
\thispagestyle{empty}

\begin{abstract}
The increasing popularity of online services has made Internet Traffic Classification a critical field of study. However, the rapid development of internet protocols and encryption limits usable data availability. This paper addresses the challenges of classifying encrypted internet traffic, focusing on the scarcity of open-source datasets and limitations of existing ones. We propose two Data Augmentation (DA) techniques to synthetically generate data based on real samples: \textit{Average} augmentation and \textit{MTU} augmentation. Both augmentations are aimed to improve the performance of the classifier, each from a different perspective: The \textit{Average} augmentation aims to increase dataset size by generating new synthetic samples, while the \textit{MTU} augmentation enhances classifier robustness to varying Maximum Transmission Units (MTUs). Our experiments, conducted on two well-known academic datasets and a commercial dataset, demonstrate the effectiveness of these approaches in improving model performance and mitigating constraints associated with limited and homogeneous datasets. Our findings underscore the potential of data augmentation in addressing the challenges of modern internet traffic classification. Specifically, we show that our augmentation techniques significantly enhance encrypted traffic classification models. This improvement can positively impact user Quality of Experience (QoE) by more accurately classifying traffic as video streaming (e.g., YouTube) or chat (e.g., Google Chat). Additionally, it can enhance Quality of Service (QoS) for file downloading activities (e.g., Google Docs). 
\end{abstract}

\begin{IEEEkeywords}
Encrypted Networks Classification, Data Augmentation, QoS/QoE
\end{IEEEkeywords}

\maketitle

\section{Introduction}
\label{sec:intro}
For the past decade, Internet Traffic Classification (ITC) has been extensively researched in a wide range of fields\cite{CERQUITELLI2023109807}. Some notable fields include cybersecurity  \cite{Papadogiannaki2021survey}, Quality of Experience (QoE)\cite{Shapira2021FlowPic:}, and 
Furthermore, with the rapid advancement in encryption, some of these datasets are losing effectiveness. Second, existing datasets are limited, with only a few classes in each dataset. Thus, models are limited in their generalized use \cite{Jorgensen2022Extensible, Rezaei2019How, loh2022youtube}. Finally, recording new data is expensive, and the protocol transition leads to data drift, which leads to the reduced effectiveness of recorded data, making the process more expensive.    

Few-Shot Learning (FSL) provides several approaches to address the challenge of insufficient data, with Data Augmentation (DA) being one of the most effective. DA encompasses various techniques to expand dataset size by transforming existing data samples. It serves two primary purposes: \textbf{I.} Generating similar yet distinct data samples through transformations of existing data. \textbf{II.} Creating new data distributions to enhance model generalization and adaptability to a wider range of inputs. By introducing new data variants, DA allows the model to learn more effectively and improve its performance. Moreover, it enables the model to represent data patterns not present in the original dataset, thus broadening its applicability.    

In this paper, we present two augmentation techniques, each addressing a specific DA goal. The first technique, which we call \textit{Average} augmentation, is a structural augmentation method. It takes a set of $m$ data flows as input and computes a new flow where the value (i.e., packet size) at each timestamp is the average of the corresponding values from the $m$ input flows with the benefit of creating a new sample in compare to other DA techniques that create a new representation for the same sample.
The second technique, \textit{MTU} augmentation, is a network-based approach. This method generates synthetic variants of each input flow by varying the Maximum Transmission Unit (MTU). For instance, the image construction method in \cite{horowicz2022few} assumes a fixed MTU of 1500 bytes. Such assumptions can limit model performance, because a small configuration change of the MTU can make the model unuseful, create opportunities for detection evasion, and expose systems to security vulnerabilities, as demonstrated in Section~\ref{sec:results}. Our \textit{MTU} augmentation addresses the second goal of DA: generating data that doesn't exist in the original dataset.
These augmentation techniques can significantly enhance encrypted traffic classification models. The improvements can positively impact user Quality of Experience (QoE) by enabling more accurate traffic classification. For example, correctly identifying video streaming (e.g., YouTube) or chat (e.g., Google Chat) traffic can lead to better bandwidth allocation. Furthermore, it can improve Quality of Service (QoS) for file downloading activities (e.g., Google Docs).

The rest of the paper is structured as follows: First, Section~\ref{sec:RW} provides a review of related work to contextualize the research. Section \ref{sec:Solution} details our solution, including augmentation methods and algorithms. The experimental design section explains the research process, datasets, evaluation metrics, preprocessing, and experiments. The results section analyzes experimental outcomes and the solution's impact on datasets. Finally, the discussion and conclusion summarize the key findings and their significance. 


\section{Related Work}\label{sec:RW}
With the growing popularity and efficiency of DL models, Internet Traffic Classification problems are migrating and using DL techniques for their versatility and flexibility \cite{Shapira2021FlowPic:, Muehlstein2020Robust, Salman2020review, Papadogiannaki2021survey, Roy2022Fast, finamore2023replication}.
Aceto et al. \cite{Aceto2021DISTILLER:} developed a DL model to classify encrypted traffic using multiple data modalities, such as packet headers, payloads, and timestamps. Bader et al. \cite{bader2022maldist} later extended this model to classify malicious traffic by adding a new modality of data, namely the content of the packets. Marin et al. \cite{marin2021deepmal} bypassed the traditional approach of extracting features from raw bytestream data using a DL network to classify traffic directly from the raw data.

A widely known rule of thumb is 'more data train better DL models' \cite{halevy2009unreasonable}. But it's not always easy to get access to a large dataset, and with the context of ITC the ability to achieve a large quantity of data is decreasing. DA-based solutions have been introduced to solve this problem. One of the first applications of DA was used in LeNet-5 \cite{lecun1998gradient}, where they distorted images using various image transformations. Krizhevsky et al. \cite{krizhevsky2012imagenet} present two distinct DA techniques.
Shorten and Khoshgoftaar~\cite{shorten2019survey}, as well as  Cubuk et al.~\cite{cubuk2020randaugment}, investigate data augmentation for image-based deep learning models. While~\cite{shorten2019survey} provides a comprehensive survey of traditional techniques,~\cite{cubuk2020randaugment} propose RandAugment, a novel automated method with a reduced search space. Notably, these works focus on the impact of augmentation on model performance metrics like accuracy, generalization, and robustness, rather than the correctness of individual augmented samples. The evaluation is typically done by comparing the performance of models trained with and without augmented data.
Inoue \cite{inoue2018data} suggests the DA solution SamplePairing, where he uses two images from the dataset to generate a new image by averaging the intensity of the pixels at each RGB channel. This is closely related to the Average augmentation we offer in this paper, with a few differences. First, our solution is dedicated to the ITC problem rather than image classification; thus, our solution is not limited to images but rather is employed in ITC situations. Second, instead of randomly choosing two images that can be from different classes, we combine only images from the same class. Lastly, the training process is different. For example,  Inoue \cite{inoue2018data} used a more convoluted and complex than the simple classic Fine-Tuning process used in this paper. Focusing on ITC DA solutions, Horowicz et al.~\cite{horowicz2022few}
proposed a set of techniques to increase the size of their data set by adding variations. Some of these techniques were borrowed from image augmentation, such as rotation, color jitter, and horizontal flip. Others were designed explicitly for internet traffic, simulating changes that occur on the internet, such as changing the round-trip time (RTT), shifting the time stamps, and dropping packets. Our paper addresses their solutions and presents more solutions to expand the DA toolkit by investigating different aspects of internet traffic.

\section{Our Solution}
\label{sec:Solution}
This paper introduces two novel data augmentation techniques: \textit{Average} and \textit{MTU}. Before delving into these innovations, we will briefly review existing data augmentation methods in computer vision to provide context. Subsequently, we will present our two proposed augmentations.
First, we introduce the \textit{Average} augmentation technique. This method aims to expand the dataset by generating new samples through the averaging of two or more existing samples within the same class. This approach increases the quantity and diversity of training data available to the model.
Second, we propose the \textit{MTU} augmentation technique. This method generates samples that simulate various MTU values, thereby exposing the model to a wider range of network conditions. By doing so, we aim to enhance the model's ability to generalize across different network environments.
Both techniques are designed to address specific challenges in the field of ITC, and their implementation and impact will be discussed in detail in the following sections.
To facilitate reproducibility, we have made the complete source code for our data augmentation process publicly available in a comprehensive GitHub repository~\cite{ITCRepo}.

\subsection{well-known Computer-vision DA}
\label{sec:cv_basic_aug}
In the field of computer vision, DA techniques enhance model robustness by increasing training data diversity. Common DA methods include geometric transformations (such as rotation, flipping, cropping, and translation), color manipulations (like color jittering), and advanced techniques (such as cutout), all of which expose the model to a wider range of visual variations. Each technique has distinct effects: image rotation maintains content but can alter packet size distribution in ITC, making it less suitable unless limited to small angles. Horizontal flipping works well for periodic signals in applications like video and VoIP. Low-intensity color jitter simulates minor packet variations, but excessive noise distorts size distributions, harming model performance. Cropping, resizing, vertical flipping, translation, and cutout significantly alter original network traffic representation, disrupting class characteristics and producing unrealistic traffic patterns. While these augmentations benefit traditional image datasets, their use in network traffic data needs careful consideration to prevent degradation of model accuracy and reliability. To address these issues, we propose two new augmentation techniques that are more suitable for network traffic data.

\subsection{Average augmentation}
\label{sec:solution_aug_avg}
\begin{figure}[htbp]
    \input{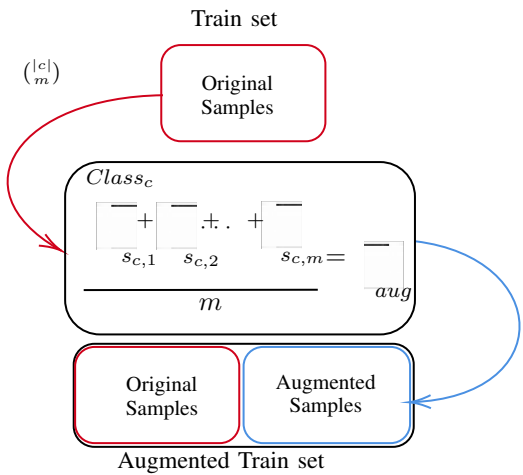}
    \caption{This workflow illustrates how new samples are extracted for each subgroup of size $m$ for any class $C$}
    \label{fig:avg_diag}
\end{figure}

We aim to increase the dataset size so that each class will have more samples from the same data distribution.
To do so, we transform the existing data to change each sample enough to count as new data yet still count as the same class. In ITC, we look at transformations that keep the samples as internet flows of the same application. Specifically, our solution takes $m$ flows from each class and calculates their average. Since each new flow is generated by combining only flows from the same class, we hypothesize that the new flow will behave similarly to the samples from which it is made up and thus is labeled accordingly. Unlike other augmentation techniques that modify existing samples, this approach introduces genuinely new samples to the dataset. To allow our solution to be as general as possible, we look at the most basic information in internet traffic: Time and Size. Therefore, to calculate the average using only these two features,
the \textit{Average} augmentation takes $m$ flows of the same class and calculates the average size at each time stamp as can be seen in Algorithm \ref{alg:Average_aug} and represented in Figure~\ref{fig:avg_diag}.

\begin{algorithm}[h]
\caption{Average Augmentation}
\label{alg:Average_aug}
\begin{algorithmic}[1]
\REQUIRE flows, m
\ENSURE augmentedData
\STATE augmentedData $\gets$ []
\FOR{each combination in combinations(flows, m)}
    \STATE flow $\gets$ combination[0]
    \FOR{i from 1 to m-1}
        \STATE flow $\gets$ flow + combination[i]
    \ENDFOR
    \STATE flow $\gets$ flow / m
    \STATE augmentedData.append(flow)
\ENDFOR
\RETURN augmentedData
\end{algorithmic}
\end{algorithm}

Doing so increases the dataset by $\binom{N}{m}$, and looking at the range of m: $ 1 \le m \le N$, we can increase the dataset size by a factor of $2^N$. However, as $m$ increases, the difference between new flows decreases since they are built more and more from the same flows, and the differentiating flows have less weight on the new flow. Furthermore, even for small values of $m$, the data becomes over-saturated by the new data, causing the original data to be much less meaningful to the model. Moreover, as this augmentation increment of each class depends on the original class size, this further magnifies the difference between imbalanced classes. Consequently, the classic use of augmentations, adding the generated data to the train data,  might not result in better performance. Hence, we offer a different solution. 

Since the data generated by the \textit{Average} augmentation is similar to the real data, we use the generated data for the Fine-Tuning process, which works as follows:
\begin{enumerate}
    \item \textbf{Pretrain the model on a large similar dataset}. The first layers of a DL model are used for feature extraction. When training a model on a similar dataset, the feature extraction can be transferred seemingly to the real data. As average augmentation generates a similar dataset, the generated data can be used to pretrain the model.
    \item \textbf{Remove the last layer(s)}.
    It is common for the last layer(s) to be used by the model to fit its target (in our case, application type classification). When changing the model's target or data, this layer(s) must be removed as the target has changed. \textit{Average} augmentation keeps the same class (i.e., the same target), meaning that most of the model should work the same for both the generated and the real data, thus requiring only the last softmax layer to be removed.
    \item \textbf{Add new layers}. New layers are added to compensate for the removed layer(s) in the previous step. As described, the model's old data and target are similar to the new data and target, only with more data and a greater imbalance (if that exists). Three new layers are added to give the real data more weight: two fully-connected (FC) layers and a softmax activation layer.
    \item \textbf{Freeze the old layers and train the new model on the target dataset.} To benefit from the old data, the old layers are "frozen", meaning that while training, the layers, and weights are not changed, only used. This step trains the new layers to fit the output given by the old layers on the real data.  
    \item \textbf{Unfreeze the layers, reduce the learning rate, and train the new model on the target dataset.} Finally, this step aims to fine-tune the weights of the entire model to fit the real data. The learning rate reduction is meant to keep the old model mostly the same with only minor changes, multiplying the learning rate by 0.1 to achieve this goal. As we described, the new data generated by the \textit{Average} augmentation does not suit the normal data augmentation process. Yet, as the data size is significantly increased and the new data is closely related to the real data, both in samples and target, we can use the new data for fine-tuning.
\end{enumerate}

\subsection{MTU augmentation}
\label{sec:solution_aug_mtu}

Maximum Transmission Unit (MTU) is the largest packet or frame size, specified in octets (bytes), that can be sent in a packet-based or frame-based network. The Transmission Control Protocol (TCP) uses the MTU to determine the maximum size of each segment in any transmission. One motivation for using different MTU values, besides network limitations, is to potentially bypass mitigation techniques used in malware traffic classification~\cite{feng2022pmtud}. 
The distribution of packet sizes in different datasets is also a motivating factor for considering different MTU sizes. For example, as can be seen in Table~\ref{tab:max_size_dist}, the maximum packet size in QUIC Davis data is distributed differently. Specifically, no packet is of size 1500 (i.e., the default MTU). As indicated by this table, many samples (almost 30\%) have a maximum packet size of less than 1300 bytes, which significantly impacts the model's ability to identify the correct class when the MTU size is changed.
Our goal is to generalize our model. However, assuming a fixed data MTU (most commonly 1500 bytes) limits the model's performance to data that fits this assumption (Note that empirical results from our experiments demonstrate this limitation in Section \ref{sec:results}). Therefore, simulating different MTUs will expand the model's target space, making it more suitable for generalized use.

\begin{table}[htbp]
    \centering
    \caption{Distribution of maximum packet size in each sample of QUIC Davis dataset}
    \renewcommand{\arraystretch}{1.3} 
    \setlength{\tabcolsep}{10pt} 
    \begin{tabular}{ |c|c|c| }
        \hline
        \textbf{max packet size} & \textbf{number of samples} & \textbf{percentage} \\ 
        \hline
        1412 & 4,021 & 60.84\% \\ 
        1294 & 1,952 & 29.54\% \\ 
        1392 & 308 & 4.66\% \\
        1434 & 294 & 4.45\% \\
        1484 & 34 & 0.51\% \\
        Other & 49 & 0.74\% \\
        \hline
    \end{tabular}
    \label{tab:max_size_dist}
\end{table}

\begin{algorithm}
\caption{MTU Augmentation}
\label{alg:MTU_aug}
\begin{algorithmic}[1]
\REQUIRE flow, MTU
\ENSURE augmentedFlow
\STATE Initialize augmentedFlow with the values of flow
\STATE Add the sum of the values from index MTU onwards in flow to the MTU index of augmentedFlow
\STATE Add the values from MTU index onwards in flow to the indices from the start to 1500-MTU in augmentedFlow
\STATE Set the values from MTU + 1 onwards in augmentedFlow to 0
\RETURN augmentedFlow
\end{algorithmic}
\end{algorithm}
{algorithm}
To simulate MTU, we represent each flow using vectors of times and sizes. This can be achieved by fragmenting any packet size greater than the new MTU into two packets with different sizes: the new MTU and (size - new MTU). Next, the fragmented packets are combined with the packets already present, as shown in Algorithm \ref {alg:MTU_aug}. As an example, suppose we configure the new MTU to be 1000. At a particular time step denoted as $i$, if there happens to be a packet with a size of 1400, our algorithm will handle it as follows: the packet is fragmented into two parts, one of 1000 and the other of 400 bytes. Subsequently, these two newly generated packets are added to the existing packets within the same time frame.

\section{Experimental Design}
Our models aim to classify different applications. To validate the effectiveness of our data augmentation strategy, we collected various datasets and trained classification models with and without augmented samples. Our results, as provided in Section \ref{sec:results}, demonstrate the promise and impact of our augmentations in enhancing the classifier's performance.

\subsection{Datasets}
Our study evaluates our system on three high-quality datasets, in contrast to many studies that use a single, standard academic dataset without preprocessing. We incorporate two well-known public datasets and one commercial dataset featuring real-world challenges from a cellular network provider, aiming to demonstrate the potential contributions of our solution comprehensively. 

The first dataset, known as \textbf{QUIC Davis} \cite{Rezaei2019How}, is an open-source dataset widely recognized in the literature. It contains labeled sessions from five Google services (Google Docs, Drive, Music, Search, and YouTube), generated by automatic tools simulating human behavior at UC Davis. The second dataset, known as\textbf{ QUIC Paris} \cite{tong2018novel}, is another well-known open-source dataset offering labeled QUIC sessions with service-specific labels (Google Chat, VoIP, YouTube, and Google Play Music). It includes corresponding PCAPs for comprehensive analysis. Independently, the data of \textbf{QUIC Davis} and \textbf{ QUIC Paris} was preprocessed and cleaned of any traffic that was not using the QUIC protocol. Additionally, any sessions with fewer than 100 packets or less than 15 seconds were removed. The last dataset, named \textbf{FLASH}, consists of real-world data captured in 2023 and is a commercial dataset  (Each dataset's sample distribution is presented in Table \ref{tab:preprocessing}).

\subsection{Evaluation Metrics}
We evaluate performance using common classification metrics, primarily focusing on Accuracy, Precision, Recall, and F1-score.  First, we define each metric using the four main outcomes of a simple binary problem: \textit{True Positive} (TP) is the number of Positive samples classified as Positive. \textit{True Negative} (TN) is the number of Negative samples classified as Negative. \textit{False Positive} (FP) is the number of Negative samples classified as Positive. \textit{False Negative} (FN) is the number of Positive samples classified as Negative. 
   
Based on these outcomes, the metrics are defined as follows: 
\begin{itemize}
    \item  \textbf{Accuracy} = $\frac{TP + TN}{TP + TN + FP + FN}$:
Represents the overall model performance, indicating the probability of correct classification regardless of the sample's true class. 
\item \textbf{Precision} = $\frac{TP}{TP + FP}$
Measures the model's trustworthiness when predicting a positive class. It represents the probability that a sample classified as positive is truly positive. 
\item \textbf{Recall} = $\frac{TP}{TP + FN}$ Also known as the true positive rate, indicates the model's ability to identify positive samples. It represents the probability that a truly positive sample will be correctly classified as positive. \item \textbf{F1-score} = $2 \cdot \frac{Precision \cdot Recall}{Precision + Recall}$ The harmonic mean of Precision and Recall. We focus on this metric as it balances both Precision and Recall, providing a comprehensive measure of model performance.
\end{itemize}

\subsection{Preprocessing}
For evaluation and comparison, we used the data representation described in \cite{Shapira2021FlowPic:, horowicz2022few}. Specifically, in \cite{horowicz2022few}, the authors introduce multiple DA techniques for ITC, with the addition that the representation is constructed only using flow's packet arrival time and packet size. To facilitate comparison, we followed the process presented in \cite{horowicz2022few} and extracted samples with the same data representation using the TCBench GitHub repository~\cite{tcbenchstack_tcbench_2023}. First, we filter all flows with less than 15 seconds captured, subsequently removing Google Music and Google Search from QUIC Davis dataset and File Transfer from QUIC Paris for lack of samples compared to the other classes in the dataset (Table \ref{tab:preprocessing}). Second, to keep the comparison equal between datasets, following \cite{Rezaei2019How}, we remove all flows with less than 100 packets captured within the first 15 seconds. Table \ref{tab:preprocessing} presents the results of the data preprocessing. \textit{All Samples} present the initial number of samples. \textit{More than 15 sec \& 100 packets} presents the number of samples that are more than 15 seconds and contain more than 100 packets. Next to the number is their percentage compared to \textit{All Samples}.

\begin{table}[h]
\centering
\caption{Sample distribution after preprocessing of used datasets}
\begin{tabular}{|l|l|r|r|}
\hline
\textbf{Dataset} & \textbf{Label} & \textbf{All Samples} & \textbf{Filtered Samples} \\
\hline
\multirow{5}{*}{QUIC Davis} 
& Google Docs   & 1,221 & 1,212 (99\%) \\
& Google Drive  & 1,634 & 1,088 (66\%) \\
& Google Music  & 591   & - \\
& Google Search & 1,915 & - \\
& YouTube       & 1,077 & 1,071 (99\%) \\
\hline
\multirow{5}{*}{QUIC Paris} 
& Google Music  & 36,866 & 230 (1\%) \\
& Google VoIP   & 57,714 & 2,346 (1\%) \\
& Google Chat   & 16,720 & 1,265 (1\%) \\
& YouTube       & 5,717  & 344 (1\%) \\
& File Transfer & 5,030  & - \\
\hline
\multirow{3}{*}{FLASH}     
& Facebook      & 2,164  & 151 (7\%) \\
& TikTok        & 9,445  & 87 (1\%) \\
& YouTube       & 15,861 & 387 (2\%) \\
\hline
\end{tabular}
\label{tab:preprocessing}
\end{table}

\subsection{Augmentation Experiments}
\label{sec:experiments}
Each evaluated dataset was divided as follows: 80\% for training (further split into 80\% for model training and 20\% for validation) and 20\% for testing. We applied data augmentation techniques only to the training portion of the dataset. The test set consisted entirely of original, unaugmented samples.

We hypothesize that enhancing the train set with augmented samples will improve the model's performance on the same neutral test set. To validate our hypothesis, we trained a model using the original dataset and tested it only with the augmented samples. Our findings result in a perfect classification, apart from one scenario in which the model classified Google Hangouts - VoIP samples as Google Hangouts - Chat ones. We encourage future work in this direction. 

We compare three variants of the original datasets:
\textbf{(1) Original} -  the original dataset without alterations.
\textbf{(2)Average 2} - dataset generated by our \textit{Average} augmentation. \textbf{(3) MTU} - dataset generated by our \textit{MTU} augmentation.

For \textit{Average} augmentation, we fixed m = 2, and for the \textit{MTU} augmentation, for each flow, the new MTU is sampled i.i.d from (750, 1200). The rationale behind this range is keeping the augmentation simple yet meaningful, as a smaller bound than 750 will force the augmentation to split each packet into three fragments, and a higher bound than 1200 might not change the flow. 

For the \textit{Average} augmentation experiment, we trained two models and then we tested the performance of the models on the \textit{Original} test dataset:
\begin{itemize}
    \item \textbf{Original}  - a model trained only on the \textit{Original} dataset.
    
    
    
    \item \textbf{Original + Average 2} - a model trained on \textit{Average 2} dataset and fine-tuned toward the \textit{Original} generated samples. 
\end{itemize}
For the \textit{MTU} augmentation experiment, we trained an additional model: \textbf{Original + MTU}, which was trained on both the \textit{Original} and \textit{MTU}-generated samples. We conducted three tests: 
\begin{enumerate}
    \item Performance comparison of the \textit{Original} model on:
a) The \textit{Original} test dataset, and b) The \textit{Original} test dataset after MTU reduction simulation. 
\item Performance comparison on the \textit{Original} test dataset between: a) The \textit{Original} model, and b) The \textit{Original + MTU} model. 
\item Performance comparison on the MTU-reduced test dataset between: a) The \textit{Original} model, and b) The \textit{Original + MTU} model. 

We employed a LeNet-5 style architecture due to its established efficacy in image recognition tasks. This model has demonstrated powerful performance, as noted by Shapira et al.~\cite{Shapira2021FlowPic:} for the miniFlowPic representation and Horowicz et al.~\cite{horowicz2022few} which also use miniFlowPic image representation with their proposed data augmentations. Additionally, its simplicity, efficiency, and adaptability to various classification problems make it an ideal choice. Table~\ref{tab:architecture} provides a detailed description of the model architecture.

\end{enumerate}

\begin{table}[htbp]
\centering
\caption{Network Architecture and Hyperparameters - based on LeNet-5}
\begin{tabular}{|p{2.5cm}|p{2.5cm}|p{2cm}|} 
\hline
\textbf{Layer Type} & \textbf{Output Shape} & \textbf{Activation}  \\ \hline
Input & (32, 32, 1)  & None  \\ 
Conv2D & (28, 28, 6)  & ReLU  \\ 
MaxPooling2D & (14, 14, 6)  & None  \\ 
Conv2D & (10, 10, 16) & ReLU  \\ 
Dropout (0.25) & (10, 10, 16) & None  \\
MaxPooling2D & (5, 5, 16)   & None  \\ 
Flatten & (400)         & None  \\
Dense & (120)         & ReLU  \\ 
Dense & (84)          & ReLU  \\
Dropout (0.5) & (84)           & None  \\
Dense & (num\_of\_classes) & Softmax \\ \hline
\end{tabular}
\label{tab:architecture}
\end{table}

\section{Results}
\label{sec:results}
This section details the results of our experiments, starting with \textit{Average} augmentation comparing existing augmentations, and ending with \textit{MTU} augmentations.
It is important to note that we conducted a comprehensive comparison with the related work mentioned in~\ref{sec:RW} on DA techniques, finding our results to be of comparable quality. However, given our focus on expanding the types of DA suitable for ITC, we determined that a detailed comparison with other techniques is beyond the scope of this paper. Therefore, due to space constraints, we have not included these results.

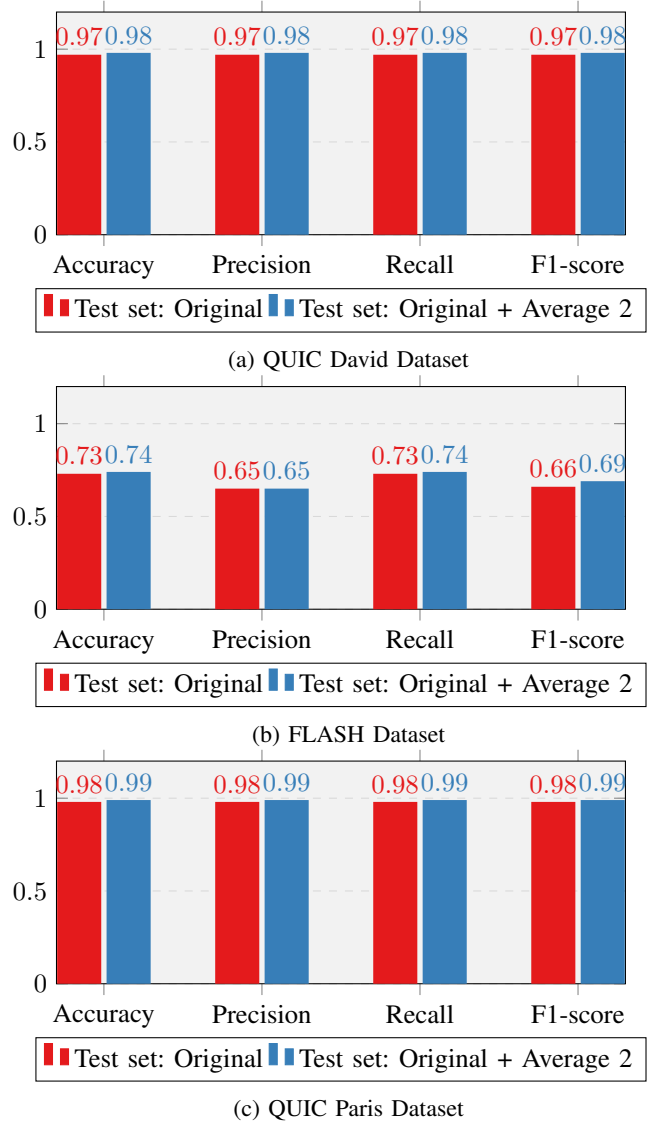
\begin{figure}[h]
    \centering
    \begin{subfigure}[t]{0.5\textwidth}
        \begin{tikzpicture}[scale=1, every node/.style={scale=1}]
\begin{axis}[
    ybar,
    bar width=.58cm,
    width=\textwidth,
    height=.5\textwidth,
    legend style={at={(0.5,-0.24)}, anchor=north,legend columns=-1},
    ylabel={},
    symbolic x coords={Accuracy,Precision,Recall,F1-score},
    xtick=data,
    nodes near coords,
    nodes near coords align={vertical},
    ymin=0,ymax=1.2,
    axis background/.style={fill=gray!10},
    grid style={dashed,gray!30},
    ymajorgrids=true,
    cycle list/Set1-4,
    every axis plot/.append style={fill,draw=none,no markers}
    ]
\addplot coordinates {(Accuracy,0.97) (Precision,0.97) (Recall,0.97) (F1-score,0.97)};
\addplot coordinates {(Accuracy,0.98) (Precision,0.98) (Recall,0.98) (F1-score,0.98)};
\legend{Test set: Original,Test set: Original + Average 2}
\end{axis}
\end{tikzpicture}
        \caption{QUIC David Dataset}
        \label{fig:DA_comparison_train_davis}
    \end{subfigure}
    \begin{subfigure}[t]{0.5\textwidth}
        \begin{tikzpicture}[scale=1, every node/.style={scale=1}]
\begin{axis}[
    ybar,
    bar width=.58cm,
    width=\textwidth,
    height=.5\textwidth,
    legend style={at={(0.5,-0.24)}, anchor=north,legend columns=-1},
    ylabel={},
    symbolic x coords={Accuracy,Precision,Recall,F1-score},
    xtick=data,
    nodes near coords,
    nodes near coords align={vertical},
    ymin=0,ymax=1.2,
    axis background/.style={fill=gray!10},
    grid style={dashed,gray!30},
    ymajorgrids=true,
    cycle list/Set1-4,
    every axis plot/.append style={fill,draw=none,no markers}
    ]
\addplot coordinates {(Accuracy,0.73) (Precision,0.65) (Recall,0.73) (F1-score,0.66)};
\addplot coordinates {(Accuracy,0.74) (Precision,0.65) (Recall,0.74) (F1-score,0.69)};
\legend{Test set: Original,Test set: Original + Average 2}
\end{axis}
\end{tikzpicture}
        \caption{FLASH Dataset}        \label{fig:DA_comparison_train_flash}
    \end{subfigure}
    \begin{subfigure}[t]{0.5\textwidth}
        \begin{tikzpicture}[scale=1, every node/.style={scale=1}]
\begin{axis}[
    ybar,
    bar width=.58cm,
    width=\textwidth,
    height=.5\textwidth,
    legend style={at={(0.5,-0.24)}, anchor=north,legend columns=-1},
    ylabel={},
    symbolic x coords={Accuracy,Precision,Recall,F1-score},
    xtick=data,
    nodes near coords,
    nodes near coords align={vertical},
    ymin=0,ymax=1.2,
    axis background/.style={fill=gray!10},
    grid style={dashed,gray!30},
    ymajorgrids=true,
    cycle list/Set1-4,
    every axis plot/.append style={fill,draw=none,no markers}
    ]
\addplot coordinates {(Accuracy,0.98) (Precision,0.98) (Recall,0.98) (F1-score,0.98)};
\addplot coordinates {(Accuracy,0.99) (Precision,0.99) (Recall,0.99) (F1-score,0.99)};
\legend{Test set: Original,Test set: Original + Average 2}
\end{axis}
\end{tikzpicture}
        \caption{QUIC Paris Dataset}
        \label{fig:DA_comparison_train_paris}
    \end{subfigure}
    \caption{A comparison between training model with and without our average augmentation. In all tests, the test set is composed of the original samples. The \textit{Model: Original} model was solely trained on the original dataset, whereas the second model was trained on both the original dataset and the average augmentation.}
    \label{fig:DA_comparison}
\end{figure}
\begin{figure}[htbp]
\begin{subfigure}[t]{1\columnwidth}
        \centering
            \begin{tikzpicture}[scale=1, every node/.style={scale=1}]
\begin{axis}[
    ybar,
    bar width=.5cm,
    width=\textwidth,
    height=.5\textwidth,
    legend style={at={(0.5,-0.24)}, anchor=north,legend columns=-1},
    ylabel={},
    symbolic x coords={Accuracy,Precision,Recall,F1-score},
    xtick=data,
    nodes near coords,
    nodes near coords align={vertical},
    ymin=0,ymax=1.2,
    axis background/.style={fill=gray!10},
    grid style={dashed,gray!30},
    ymajorgrids=true,
    cycle list/Set1-4,
    every axis plot/.append style={fill,draw=none,no markers}
    ]
\addplot coordinates {(Accuracy,0.97) (Precision,0.97) (Recall,0.97) (F1-score,0.97)};
\addplot coordinates {(Accuracy,0.52) (Precision,0.44) (Recall,0.52) (F1-score,0.43)};
\legend{Test set: Original,Test set: MTU}
\end{axis}
\end{tikzpicture}
        \caption{QUIC Davis Dataset}
        \label{fig:MTU_opening_train_davis}
    \end{subfigure}
    
    \begin{subfigure}[t]{0.5\textwidth}
        \centering
            \begin{tikzpicture}[scale=1, every node/.style={scale=1}]
\begin{axis}[
    ybar,
    bar width=.5cm,
    width=\textwidth,
    height=.5\textwidth,
    legend style={at={(0.5,-0.24)}, anchor=north,legend columns=-1},
    ylabel={},
    symbolic x coords={Accuracy,Precision,Recall,F1-score},
    xtick=data,
    nodes near coords,
    nodes near coords align={vertical},
    ymin=0,ymax=1.2,
    axis background/.style={fill=gray!10},
    grid style={dashed,gray!30},
    ymajorgrids=true,
    cycle list/Set1-4,
    every axis plot/.append style={fill,draw=none,no markers}
    ]
\addplot coordinates {(Accuracy,0.73) (Precision,0.65) (Recall,0.73) (F1-score,0.66)};
\addplot coordinates {(Accuracy,0.31) (Precision,0.59) (Recall,0.31) (F1-score,0.19)};
\legend{Test set: Original,Test set: MTU}
\end{axis}
\end{tikzpicture}
            \caption{FLASH Dataset}
        \label{fig:MTU_opening_train_flash}
    \end{subfigure}
    
    \begin{subfigure}[t]{0.5\textwidth}
        \centering
            \begin{tikzpicture}[scale=1, every node/.style={scale=1}]
\begin{axis}[
    ybar,
    bar width=.5cm,
    width=\textwidth,
    height=.5\textwidth,
    legend style={at={(0.5,-0.24)}, anchor=north,legend columns=-1},
    ylabel={},
    symbolic x coords={Accuracy,Precision,Recall,F1-score},
    xtick=data,
    nodes near coords,
    nodes near coords align={vertical},
    ymin=0,ymax=1.2,
    axis background/.style={fill=gray!10},
    grid style={dashed,gray!30},
    ymajorgrids=true,
    cycle list/Set1-4,
    every axis plot/.append style={fill,draw=none,no markers}
    ]
\addplot coordinates {(Accuracy,0.99) (Precision,0.99) (Recall,0.99) (F1-score,0.99)};
\addplot coordinates {(Accuracy,0.83) (Precision,0.81) (Recall,0.83) (F1-score,0.8)};
\legend{Test set: Original,Test set: MTU}
\end{axis}
\end{tikzpicture}
            \caption{QUIC Paris Dataset}
        \label{fig:MTU_opening_train_paris}
    \end{subfigure}
\caption{Comparing performance decline with MTU lower than 1500. In both cases, we used the same model trained on the Original dataset and changed the test dataset. \textbf{Test: Original} is the Original test samples, and \textbf{Test: MTU} is the simulated reduced MTU samples.}
\label{fig:MTU_opening}
\end{figure}
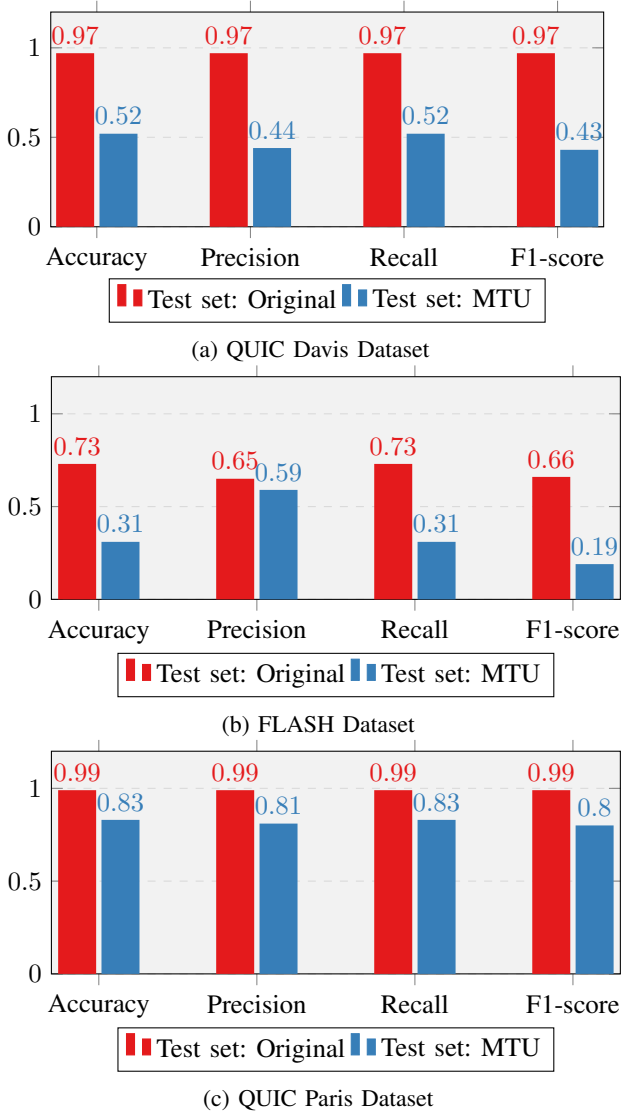

\subsection{Average Augmentation}
A comparison of network augmentations is presented in Figure \ref{fig:DA_comparison}. 
Since QUIC Davis and QUIC Paris datasets are balanced \textit{Average}, augmentation did not change the balance of the classes, thus achieving great results across the board with a 1\% increase in all metrics. However, with the high results in the \textit{Original} model alone, a more realistic and representative dataset of ITC needs to be explored. Figure \ref{fig:DA_comparison_train_flash} displays more extreme results for both the  \textit{Original} and \textit{Original + Average} models, indicating the issue of data imbalance and lack of data, as FLASH data has significantly fewer samples and greater class imbalance. The \textit{Average} augmentation surpassed \textit{Original} in all metrics apart from precision, \textit{Average} acted similarly to \textit{Original}, with accuracy and recall being greater than 0.7 and Precision and F1-Score lower than 0.7. 
While results may exhibit some variability, the effectiveness of the \textit{Average} augmentation is comparable to other presented augmentation techniques, making it a viable option for standalone use or in combination with other augmentation methods.

\subsection{MTU Augmentation}

In this section, we present three experiments: the over-reliance on MTU being 1500, the lost performance when using \textit{MTU} augmentation, and the performance improvement when using \textit{MTU} augmentation on data that does not follow a fixed MTU. First, in Figure \ref{fig:MTU_opening}, we 
compare the performance of the \textit{Original} model on two datasets - the \textit{Original} test dataset, a modified version of the same dataset that simulates a smaller MTU (e.g., change by an attacker). 

The results provide insight into the model's performance when the MTU is altered. It is evident that the performance decreased significantly, with a 49\% drop in QUIC Davis performance, a 16\% drop in QUIC Paris, and a 34\% drop in FLASH performance. This suggests that the model is vulnerable and relies heavily on an MTU of 1500. This drop is made more significant when looking at the classes. Most classes in the datasets require a large MTU, especially YouTube, as this is a video streaming service. Thus, changing the MTU changes the data almost completely. This effect can also be seen in the following experiment.

Figure~\ref{fig:MTU_performance_train_flash} depicts the performance drop when the \textit{MTU} augmentation data is added to the train dataset. This reduction is due to the new data acting differently from the original data, which confuses the model, causing performance reduction. This result reinforces the idea that different MTUs can create an opening for detection evasion in the model. 
    
\begin{figure*}[!]
    \begin{subfigure}[t]{0.45\textwidth}
        \centering
            \begin{tikzpicture}[scale=1, every node/.style={scale=1}]
\begin{axis}[
    ybar,
    bar width=.5cm,
    width=\textwidth,
    height=.5\textwidth,
    legend style={at={(0.5,-0.24)}, anchor=north,legend columns=-1},
    ylabel={},
    symbolic x coords={Accuracy,Precision,Recall,F1-score},
    xtick=data,
    nodes near coords,
    nodes near coords align={vertical},
    ymin=0,ymax=1.2,
    axis background/.style={fill=gray!10},
    grid style={dashed,gray!30},
    ymajorgrids=true,
    cycle list/Set1-4,
    every axis plot/.append style={fill,draw=none,no markers}
    ]
\addplot coordinates {(Accuracy,0.97) (Precision,0.97) (Recall,0.97) (F1-score,0.97)};
\addplot coordinates {(Accuracy,0.97) (Precision,0.97) (Recall,0.97) (F1-score,0.97)};
\legend{Test set: Original, Test set: Original + MTU}
\end{axis}
\end{tikzpicture}
        \caption{Tested without MTU augmentation}
        \label{fig:MTU_performance_train_davis}
    \end{subfigure}
    \begin{subfigure}[t]{0.45\textwidth}
        \centering
            \begin{tikzpicture}[scale=1, every node/.style={scale=1}]
\begin{axis}[
    ybar,
    bar width=.5cm,
    width=\textwidth,
    height=.5\textwidth,
    legend style={at={(0.5,-0.24)}, anchor=north,legend columns=-1},
    ylabel={},
    symbolic x coords={Accuracy,Precision,Recall,F1-score},
    xtick=data,
    nodes near coords,
    nodes near coords align={vertical},
    ymin=0,ymax=1.2,
    axis background/.style={fill=gray!10},
    grid style={dashed,gray!30},
    ymajorgrids=true,
    cycle list/Set1-4,
    every axis plot/.append style={fill,draw=none,no markers}
    ]
\addplot coordinates {(Accuracy,0.52) (Precision,0.44) (Recall,0.52) (F1-score,0.43)};
\addplot coordinates {(Accuracy,0.63) (Precision,0.79) (Recall,0.67) (F1-score,0.64)};
\legend{Test set: Original, Test set: Original + MTU}
\end{axis}
\end{tikzpicture}
        \caption{Tested on test set with simulated reduced MTU}
        \label{fig:MTU_aug_MTU_test_train_davis}
    \end{subfigure}
    \caption{QUIC Davis Dataset}
    \label{fig:davis_testset_with_without_mtu}
\end{figure*}
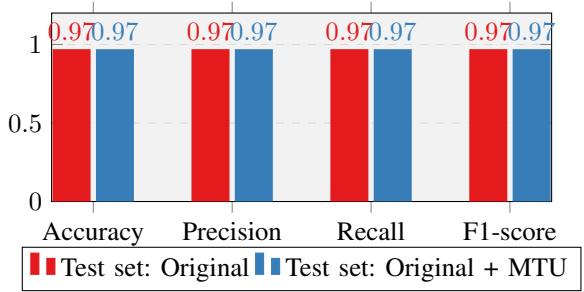
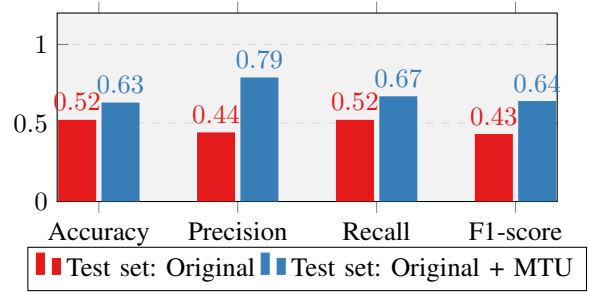
\begin{figure*}
    \begin{subfigure}[t]{0.45\textwidth}
        \centering
            \begin{tikzpicture}[scale=1, every node/.style={scale=1}]
\begin{axis}[
    ybar,
    bar width=.5cm,
    width=\textwidth,
    height=.5\textwidth,
    legend style={at={(0.5,-0.24)}, anchor=north,legend columns=-1},
    ylabel={},
    symbolic x coords={Accuracy,Precision,Recall,F1-score},
    xtick=data,
    nodes near coords,
    nodes near coords align={vertical},
    ymin=0,ymax=1.2,
    axis background/.style={fill=gray!10},
    grid style={dashed,gray!30},
    ymajorgrids=true,
    cycle list/Set1-4,
    every axis plot/.append style={fill,draw=none,no markers}
    ]
\addplot coordinates {(Accuracy,0.73) (Precision,0.65) (Recall,0.73) (F1-score,0.66)};
\addplot coordinates {(Accuracy,0.61) (Precision,0.51) (Recall,0.61) (F1-score,0.5)};
\legend{Test set: Original, Test set: Original + MTU}
\end{axis}
\end{tikzpicture}
        \caption{Tested without MTU augmentation}
        \label{fig:MTU_performance_train_flash}
    \end{subfigure}
    \begin{subfigure}[t]{0.45\textwidth}
        \centering
            \begin{tikzpicture}[scale=1, every node/.style={scale=1}]
\begin{axis}[
    ybar,
    bar width=.5cm,
    width=\textwidth,
    height=.5\textwidth,
    legend style={at={(0.5,-0.24)}, anchor=north,legend columns=-1},
    ylabel={},
    symbolic x coords={Accuracy,Precision,Recall,F1-score},
    xtick=data,
    nodes near coords,
    nodes near coords align={vertical},
    ymin=0,ymax=1.2,
    axis background/.style={fill=gray!10},
    grid style={dashed,gray!30},
    ymajorgrids=true,
    cycle list/Set1-4,
    every axis plot/.append style={fill,draw=none,no markers}
    ]
\addplot coordinates {(Accuracy,0.31) (Precision,0.59) (Recall,0.31) (F1-score,0.19)};
\addplot coordinates {(Accuracy,0.57) (Precision,0.46) (Recall,0.57) (F1-score,0.48)};
\legend{Test set: Original, Test set: Original + MTU}
\end{axis}
\end{tikzpicture}
        \caption{Tested on test set with simulated reduced MTU}
        \label{fig:MTU_aug_MTU_test_train_flash}
    \end{subfigure}
    
    \caption{Flash Dataset}
    \label{fig:flash_testset_with_without_mtu}
\end{figure*}
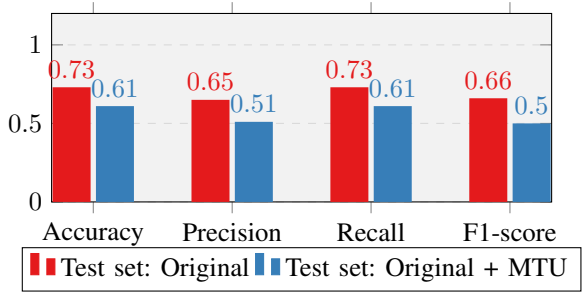
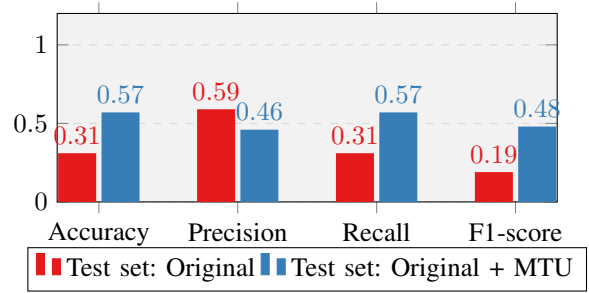
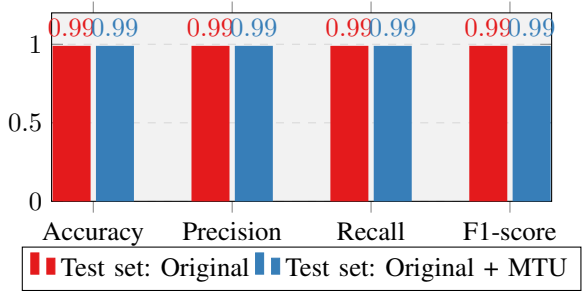
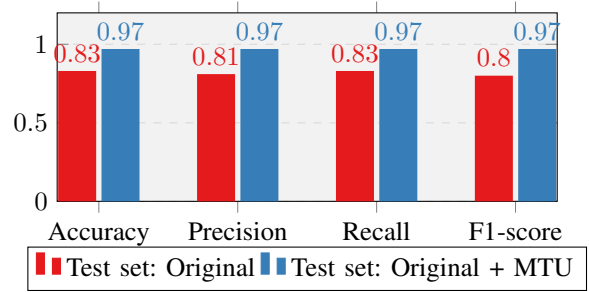
\begin{figure*}
    \begin{subfigure}[t]{0.45\textwidth}
        \centering
            \begin{tikzpicture}[scale=1, every node/.style={scale=1}]
\begin{axis}[
    ybar,
    bar width=.5cm,
    width=\textwidth,
    height=.5\textwidth,
    legend style={at={(0.5,-0.24)}, anchor=north,legend columns=-1},
    ylabel={},
    symbolic x coords={Accuracy,Precision,Recall,F1-score},
    xtick=data,
    nodes near coords,
    nodes near coords align={vertical},
    ymin=0,ymax=1.2,
    axis background/.style={fill=gray!10},
    grid style={dashed,gray!30},
    ymajorgrids=true,
    cycle list/Set1-4,
    every axis plot/.append style={fill,draw=none,no markers}
    ]
\addplot coordinates {(Accuracy,0.99) (Precision,0.99) (Recall,0.99) (F1-score,0.99)};
\addplot coordinates {(Accuracy,0.99) (Precision,0.99) (Recall,0.99) (F1-score,0.99)};
\legend{Test set: Original, Test set: Original + MTU}
\end{axis}
\end{tikzpicture}
        \caption{Tested without MTU augmentation}
        \label{fig:MTU_performance_train_paris}
    \end{subfigure}
    \begin{subfigure}[t]{0.45\textwidth}
        \centering
            \begin{tikzpicture}[scale=1, every node/.style={scale=1}]
\begin{axis}[
    ybar,
    bar width=.5cm,
    width=\textwidth,
    height=.5\textwidth,
    legend style={at={(0.5,-0.24)}, anchor=north,legend columns=-1},
    ylabel={},
    symbolic x coords={Accuracy,Precision,Recall,F1-score},
    xtick=data,
    nodes near coords,
    nodes near coords align={vertical},
    ymin=0,ymax=1.2,
    axis background/.style={fill=gray!10},
    grid style={dashed,gray!30},
    ymajorgrids=true,
    cycle list/Set1-4,
    every axis plot/.append style={fill,draw=none,no markers}
    ]
\addplot coordinates {(Accuracy,0.83) (Precision,0.81) (Recall,0.83) (F1-score,0.8)};
\addplot coordinates {(Accuracy,0.97) (Precision,0.97) (Recall,0.97) (F1-score,0.97)};
\legend{Test set: Original, Test set: Original + MTU}
\end{axis}
\end{tikzpicture}
        \caption{Tested on test set with simulated reduced MTU}
        \label{fig:MTU_aug_MTU_test_train_paris}
    \end{subfigure}
    
    \caption{QUIC Paris Dataset}
    \label{fig:paris_testset_with_without_mtu}
\end{figure*}

Figures \ref{fig:MTU_performance_train_davis} and \ref{fig:MTU_performance_train_paris} show that for QUIC Davis and QUIC Paris datasets, the performance did not change. While somehow contradicting the earlier statements, this can be explained with context. Davis and Paris QUIC datasets have very distinct classes and a relatively high number of samples. Thus, we get a high-performing model regardless of augmentations. Therefore, the \textit{MTU} data added to the train data did not change the performance as there is enough data to cover the changes made by the \textit{MTU} augmentation. Once again, this forces us to look at a more real situation with a much more realistic dataset. 

Figure \ref{fig:MTU_performance_train_flash} shows the same outcome but with the FLASH dataset. For this dataset, the performance was reduced by an average of 13\%. This indicates the confusion that the \textit{MTU} augmentation causes to the model while having the same number of train samples from both \textit{Original} data and \textit{MTU} data. 13\% indicates the weight the \textit{MTU} data changes the model while being the same size as \textit{Original}. However, the \textit{MTU} augmentation makes up for it by increasing the performance on \textit{Original} test dataset that simulated a reduced MTU.


    


Finally, Figures \ref{fig:MTU_aug_MTU_test_train_davis}, \ref{fig:MTU_aug_MTU_test_train_flash}, and \ref{fig:MTU_aug_MTU_test_train_paris} compares the different performance of training the model on \textit{Original} train set and training the model on \textit{Original + MTU} train set on the test set that was changed to simulate a smaller MTU. The addition of \textit{MTU} augmentation increased the f1-score by 21\% on the QUIC Davis dataset, by 29\% on the FLASH dataset, and by 13\% on the QUIC Paris dataset. There was no performance loss in QUIC Davis and QUIC Paris datasets and 13\% loss in FLASH precision performance. 

\section{Discussion and Conclusions}
This paper introduces two novel augmentation techniques for internet traffic classification, each accomplishing a different goal. \textit{Average} augmentation enlarges the dataset size. Using fine-tuning, this augmentation improves the model classification. 
\textit{MTU} augmentation simulates a reduced MTU for each flow in the dataset. With the assumption that the MTU is fixed (i.e. 1500), inputs with different MTUs can be classified incorrectly, making the model vulnerable to attacks. This vulnerability arises because the model begins to rely on an outside factor. Thus, this augmentation generalizes and robusts the model by removing the assumption of a fixed MTU in the train data.

Experimenting with both a well-known dataset and an industry dataset. The results demonstrate the improvement using \textit{Average} augmentation, the model's over-reliance on a fixed MTU, and the trade-off of training the model with different MTUs. For future research, we advise experimenting with \textit{Average} augmentation, both in other datasets and problems connected to ITC and different values of $m$. \textit{MTU} augmentation is open for both attacking models and reinforcing and generalizing models. We encourage testing the outcome of models when detecting malware with an MTU lower than 1500. 
\bibliographystyle{IEEEtran}
\bibliography{sample}

\begin{thebibliography}{10}
\providecommand{\url}[1]{#1}
\csname url@samestyle\endcsname
\providecommand{\newblock}{\relax}
\providecommand{\bibinfo}[2]{#2}
\providecommand{\BIBentrySTDinterwordspacing}{\spaceskip=0pt\relax}
\providecommand{\BIBentryALTinterwordstretchfactor}{4}
\providecommand{\BIBentryALTinterwordspacing}{\spaceskip=\fontdimen2\font plus
\BIBentryALTinterwordstretchfactor\fontdimen3\font minus \fontdimen4\font\relax}
\providecommand{\BIBforeignlanguage}[2]{{%
\expandafter\ifx\csname l@#1\endcsname\relax
\typeout{** WARNING: IEEEtran.bst: No hyphenation pattern has been}%
\typeout{** loaded for the language `#1'. Using the pattern for}%
\typeout{** the default language instead.}%
\else
\language=\csname l@#1\endcsname
\fi
#2}}
\providecommand{\BIBdecl}{\relax}
\BIBdecl

\bibitem{CERQUITELLI2023109807}
\BIBentryALTinterwordspacing
T.~Cerquitelli, M.~Meo, M.~Curado, L.~Skorin-Kapov, and E.~E. Tsiropoulou, ``Machine learning empowered computer networks,'' \emph{Computer Networks}, vol. 230, p. 109807, 2023. [Online]. Available: \url{https://www.sciencedirect.com/science/article/pii/S1389128623002529}
\BIBentrySTDinterwordspacing

\bibitem{Papadogiannaki2021survey}
E.~Papadogiannaki and S.~Ioannidis, ``A survey on encrypted network traffic analysis applications, techniques, and countermeasures,'' \emph{CSUR}, vol.~54, no.~6, p. 1–35, 2021.

\bibitem{Shapira2021FlowPic:}
T.~Shapira and Y.~Shavitt, ``Flowpic: A generic representation for encrypted traffic classification and applications identification,'' \emph{IEEE TNSM}, vol.~18, no.~2, p. 1218–1232, 2021, publisher: IEEE.

\bibitem{Jorgensen2022Extensible}
S.~Jorgensen, J.~Holodnak, J.~Dempsey, K.~de~Souza, A.~Raghunath, V.~Rivet, N.~DeMoes, A.~Alejos, and A.~Wollaber, ``Extensible machine learning for encrypted network traffic application labeling via uncertainty quantification,'' \emph{arXiv preprint arXiv:2205.05628}, 2022.

\bibitem{Rezaei2019How}
S.~Rezaei and X.~Liu, ``How to achieve high classification accuracy with just a few labels: A semi-supervised approach using sampled packets,'' in \emph{ICDM}, 2019, pp. 28--42.

\bibitem{loh2022youtube}
F.~Loh, F.~Wamser, F.~Poign{\'e}e, S.~Gei{\ss}ler, and T.~Ho{\ss}feld, ``Youtube dataset on mobile streaming for internet traffic modeling and streaming analysis,'' \emph{Scientific Data}, vol.~9, no.~1, p. 293, 2022.

\bibitem{horowicz2022few}
E.~Horowicz, T.~Shapira, and Y.~Shavitt, ``A few shots traffic classification with mini-flowpic augmentations,'' in \emph{ACM IMC}, 2022, pp. 647--654.

\bibitem{Muehlstein2020Robust}
J.~Muehlstein, Y.~Zion, O.~Pele, C.~Hajaj, R.~Dubin, and A.~Dvir, ``Robust machine learning for encrypted traffic classification,'' \emph{CoRR}, vol. abs/1603.04865, 2020.

\bibitem{Salman2020review}
O.~Salman, I.~H. Elhajj, A.~Kayssi, and A.~Chehab, ``A review on machine learning–based approaches for internet traffic classification,'' \emph{Annals of Telecommunications}, p. 1–38, 2020, publisher: Springer.

\bibitem{Roy2022Fast}
S.~Roy, T.~Shapira, and Y.~Shavitt, ``Fast and lean encrypted internet traffic classification,'' \emph{Computer Communications}, vol. 186, p. 166–173, 2022, publisher: Elsevier.

\bibitem{finamore2023replication}
A.~Finamore, C.~Wang, J.~Krolikowski, J.~M. Navarro, F.~Chen, and D.~Rossi, ``Replication: Contrastive learning and data augmentation in traffic classification using a flowpic input representation,'' in \emph{IM}, 2023, pp. 36--51.

\bibitem{Aceto2021DISTILLER:}
G.~Aceto, D.~Ciuonzo, A.~Montieri, and A.~Pescapé, ``Distiller: Encrypted traffic classification via multimodal multitask deep learning,'' \emph{Journal of Network and Computer Applications}, vol. 183, p. 102985, 2021, publisher: Elsevier.

\bibitem{bader2022maldist}
O.~Bader, A.~Lichy, C.~Hajaj, R.~Dubin, and A.~Dvir, ``Maldist: From encrypted traffic classification to malware traffic detection and classification,'' in \emph{CCNC}, 2022, pp. 527--533.

\bibitem{marin2021deepmal}
G.~Mar{\'\i}n, P.~Caasas, and G.~Capdehourat, ``{Deepmal-deep learning models for malware traffic detection and classification},'' in \emph{iDSC}.\hskip 1em plus 0.5em minus 0.4em\relax Springer, 2021, pp. 105--112.

\bibitem{halevy2009unreasonable}
A.~Halevy, P.~Norvig, and F.~Pereira, ``The unreasonable effectiveness of data,'' \emph{IEEE Intelligent Systems}, vol.~24, no.~2, pp. 8--12, 2009.

\bibitem{lecun1998gradient}
Y.~LeCun, L.~Bottou, Y.~Bengio, and P.~Haffner, ``Gradient-based learning applied to document recognition,'' \emph{Proceedings of the IEEE}, vol.~86, no.~11, pp. 2278--2324, 1998.

\bibitem{krizhevsky2012imagenet}
A.~Krizhevsky, I.~Sutskever, and G.~E. Hinton, ``Image-net classification with deep convolutional neural networks,'' \emph{Advances in Neural Information Processing Systems}, vol.~25, 2012.

\bibitem{shorten2019survey}
C.~Shorten and T.~M. Khoshgoftaar, ``A survey on image data augmentation for deep learning,'' \emph{Journal of big data}, vol.~6, no.~1, pp. 1--48, 2019.

\bibitem{cubuk2020randaugment}
E.~D. Cubuk, B.~Zoph, J.~Shlens, and Q.~V. Le, ``Randaugment: Practical automated data augmentation with a reduced search space,'' in \emph{CVPR-workshops}, 2020, pp. 702--703.

\bibitem{inoue2018data}
H.~Inoue, ``Data augmentation by pairing samples for images classification,'' \emph{arXiv preprint arXiv:1801.02929}, 2018.

\bibitem{ITCRepo}
``{ITC-Data-Augmentations GitHub Repository},'' \url{https://github.com/ArielCyber/ITC-Data-Augmentations-}.

\bibitem{feng2022pmtud}
X.~Feng, Q.~Li, K.~Sun, K.~Xu, B.~Liu, X.~Zheng, Q.~Yang, H.~Duan, and Z.~Qian, ``Pmtud is not panacea: Revisiting ip fragmentation attacks against tcp,'' in \emph{NDSS}, 2022, pp. 24--28.

\bibitem{tong2018novel}
V.~Tong, H.~A. Tran, S.~Souihi, and A.~Mellouk, ``A novel quic traffic classifier based on convolutional neural networks,'' in \emph{GLOBECOM}.\hskip 1em plus 0.5em minus 0.4em\relax IEEE, 2018, pp. 1--6.

\bibitem{tcbenchstack_tcbench_2023}
\BIBentryALTinterwordspacing
{tcbenchstack}, ``{tcbench: tcbench is a Machine Learning and Deep Learning framework to train model from traffic packet time series or other input representations.}'' 2023. [Online]. Available: \url{https://github.com/tcbenchstack/tcbench}
\BIBentrySTDinterwordspacing

\end{thebibliography}

\end{document}